# Cross-Recurrence Quantification Analysis of Categorical and Continuous Time Series: an R package

Moreno I. Coco
Faculdade de Psicologia, Universidade de Lisboa
Alameda da Universidade, Lisboa, 1649-013, Portugal
micoco@fp.ul.pt

Rick Dale
Cognitive and Information Sciences, University of California, Merced
5200 North Lake Rd., Merced, CA 95343, USA
rdale@ucmerced.edu

## Abstract

This paper describes the R package crqa to perform cross-recurrence quantification analysis of two time series of either a categorical or continuous nature. Streams of behavioral information, from eye movements to linguistic elements, unfold over time. When two people interact, such as in conversation, they often adapt to each other, leading these behavioral levels to exhibit *recurrent states*. In dialogue, for example, interlocutors adapt to each other by exchanging interactive cues: smiles, nods, gestures, choice of words, and so on. In order for us to capture closely the goings-on of dynamic interaction, and uncover the extent of coupling between two individuals, we need to quantify how much *recurrence* is taking place at these lev- els. Methods available in crqa would allow researchers in cognitive science to pose such questions as how much are two people recurrent at some level of analysis, what is the characteristic lag time for one person to maximally match another, or whether one person is leading another. First, we set the theoretical ground to un- derstand the difference between 'correlation' and 'co-visitation' when comparing two time series, using an aggregative or cross-recurrence approach. Then, we de- scribe more formally the principles of cross-recurrence, and show with the current package how to carry out analyses applying them. We end the paper by compar- ing computational efficiency, and results' consistency, of crqa R package, with the benchmark MATLAB toolbox crptoolbox. We show perfect comparability be- tween the two libraries on both levels.

## Introduction

A human being is a dynamical system. Such is an incontrovertible but vacuous claim, on its surface. However, an emphasis on dynamics of human behavior is often under-emphasized, especially in the language sciences. Consider the interaction between two human beings. Interaction is a fundamentally dynamic process, as the production and comprehension of language are incremental. Yet many studies of interaction are based on *aggregative* methods, which flatten out the temporal dimension characterizing the interaction, and rather focus on the "magnitude" of such interaction.

This aggregative approach has borne considerable fruit for some questions. For example, when two people interact they may come to mimic each other as measured by behavioral frequencies (Bargh & Chartrand, 1999), and they may utilize similar sentence structures at opportune times as discerned by careful experimental design (Haywood, Pickering, & Branigan, 2005). Many papers have shown that humans can coordinate syntactic structures (Branigan, 2007), entrain on descriptions (Brennan & Clark, 1996), spatial perspective (Schober, 1993), and so on. Indeed, this aggregative approach it has been the dominant technique in the language sciences for studying the convergence of human interlocutors (we discuss prominent exceptions later in this paper).

There is no doubt that such aggregative methods are important, and often sufficient for rendering new insights into interaction. But recent work has sought to characterize the manner in which these aggregate scores unfold. Put simply, taking aggregate measures and "unfolding them in time" offers both intriguing methods, and also new questions: Does the temporal organization of interaction show interesting patterns, beneath their aggregation? Do these patterns shed light on the mechanisms underlying human interaction? How are different behavioral measures organized in time relative to each other? What variables impact the shape of coordination between two people who are interacting?

By unfolding behavioral measures, and subjecting them to temporal analysis, we can indeed find distinct dynamics between two interacting people. For example, D. C. Richardson and Dale (2005) find that when one person is speaking to a listener, they exhibit coupled gaze patterns, but with the listener's eye movements lagged by a characteristic time of about 2 seconds. Interestingly, the lag time of any one listener predicted their comprehension; the dynamics of coupling revealed comprehension. But as two people talk bidirectionally (taking turns as speaker and listener), this lag time approaches 0 seconds, suggesting tighter coupling occurs during real-time interaction (D. Richardson, Dale, & Kirkham, 2007; Dale, Kirkham, & Richardson, 2011). And beyond just eye movements, other behavioral aspects of interaction exhibit this coupling, such as nods, gestures, and conversational moves (Louwerse, Dale, Bard, & Jeuniaux, 2012).

These basic insights were generated through what is called *cross-recurrence* methods. It is a family of techniques related to measuring the manner in which, and extent to which, two people come to exhibit similar patterns of behavior *in time*. This analysis framework was developed, and is extensively employed, in the natural sciences in such diverse domains as heart rate variability, seismology, and chemical fluctuations (see Marwan, Carmen Romano, Thiel, & Kurths, 2007; Marwan, 2008, for reviews). In psychology, it rapidly gained attention in the domain of motor control (e.g., Shockley & Turvey, 2005; Stephen, Dixon, & Isenhower, 2009), being applied to both within- and between-person dynamics, such as during precision-target tasks (Balasubramaniam, Riley, & Turvey, 2000) and even conversation (Shockley, Santana, & Fowler, 2003).

As we describe further below, the method is often referred to as a "nonlinear" technique that permits the researcher to avoid certain assumptions that linear statistics make (see Riley &

Van Orden, 2005). They also reveal system characteristics, phrased in the language of dynamical systems, permitting researchers to describe their phenomena in new and potentially interesting ways. A comprehensive review of the method can be found in Marwan et al. (2007), and an especially lucid introduction to it in Webber Jr. and Zbilut (2005). An excellent MATLAB toolbox for recurrence can be found in Marwan (2013).[1] The current toolbox is not meant to replace this one, but rather to offer researchers some of these methods on the R platform. We designed the toolbox `crqa` specifically for questions relating to human behavioral dynamics, especially in the linguistic context. Narrowing our focus here still reveals a wide array of potential questions, from eye-movement patterns to conversational moves and semantics. The package allows researchers to explore the dynamics of diverse behavioral domains.

In the next section, we motivate recurrence methods, and describe the new R package that makes it freely available to researchers. To do so, we make use of the strategy of developing highly simplified models as demonstration (cf.,  Beer, 2003), in our case to generate hypothetical data from known principles. In all demonstrations, we emphasize the value of this technique for studying linguistic interaction: finding temporal patterning between two persons as they interact. We start in an unusual but, we believe, helpful manner: by motivating the importance of unfolding aggregate measures, and showing how recurrence does this.

### Motivating Recurrence: Aggregation, Covariance, and Co-Visitation

In this section, we aim to briefly motivate cross-recurrence methods, and relate them conceptually to statistical aggregation ("atemporal" aggregation), and cross-correlation approaches. We will not articulate the formal relationships among these analyses, as they have been articulated elsewhere (see, e.g., Marwan et al., 2007; Dale, Warlaumont, & Richardson, 2011). However, there are relatively few clear comparisons of these techniques that explain where and when each would be useful. Our aim here is to use a very simple toy data model to motivate cross-recurrence methods. Aggregation and correlation scores are highly useful and easy to compute, but they are not a comprehensive characterization of two systems' relative behaviors. By focusing on the path of a system's behavior in time, there may be other indices that describe how two systems are exhibiting similar or dissimilar behavioral patterns. We hope this simple section motivates the distinction between covariance-based and 'visitation-based measures'; in the next section, we provide extensive further detail how these recurrence quantities are extracted on real experimental data, and on the theoretical insights gained by using them.

Our toy model derives from a common experimental circumstance. Imagine having a confederate (C) interact with 40 subjects (S) in the laboratory. In one condition (high), you have the confederate amplify a particular pattern of behavior, such as scratching the face or touching the foot. In another condition (low) you have them minimize such behaviors. Doing an experiment much like this, Bargh and Chartrand (1999) had confederates use non-salient and seemingly incidental behaviors to induce this behavior in a communication partner. By having a confederate engage in one or the other behavior, they can induce the participant to increase their behavior along the same dimension. Researchers aggregate the observed effect on participants (how many times the participant engages in these behaviors), and find that the rate can be amplified as a function of the confederate's behavior (high vs. low rate of target behavior).

---

[1]In fact readers may consult Marwan's excellent online resources that list other software tools as well: http://recurrence-plot.tk/programmes.php

Table 1

*A simple algorithm for producing a system (C) that drives a second system (S) for a binary time series (1 for event occurrence; 0 otherwise).*

| Variables | Algorithm |
|---|---|
| $P(X)$ = base rate of event for person X | Produce a time series for C and S events: |
| $P(X\|Y)$ = rate of event for X given Y did | |
| $P(X\|X)$ = probability of event repetition | ```
Do 1000 times
  If rand < P(C)
    C outputs event (=1)
  Else if rand < P(C|C) and C = 1
    C outputs event
  Otherwise
    C outputs no event (=0)
  If rand < P(S|C) and C = 1
    S outputs event (=1)
  Else if rand < P(S)
    S outputs event
  Else if rand < P(S|S) and S = 1
    S outputs event
  Otherwise
    S outputs no event (=0)
``` |

*Notes*: In the algorithm, C = confederate agent, S = participant agent. 20 such runs were conducted for 1000 iterations for each of conditions low $P(C)$ = .05 and high $P(C)$ = .25. Other parameters include: $P(S)$ = .05, $P(C|C)$ = $P(S|S)$ = .2, and $P(S|C)$ = .25. Parameters were chosen to bring average behavior to approximately 0.1 in the low condition. This is merely for demonstration and other parameter values would work fine.

Let us take up some purely hypothetical data for the sake of demonstration, using precisely this setup. We designed a very simple simulation of the kind just described, in which we collect data about the occurrence of a specific behavioral event, across time, between confederate and participant "agents." These simplified interactants are quite simple, and obey simple parameterized behaviors that can be subjected to the analysis described here. We use R code to specify the behavior of confederate vs. participant along some dimension in Table 1. In actual practice these data may be the occurrence of touching the face or foot (Bargh & Chartrand, 1999), looks to certain characters on a computer screen (D. C. Richardson & Dale, 2005), or an entire array of behaviors from dialogue moves to laughter events (Louwerse et al., 2012). Readers may consult detailed advice and coding schemes for discrete behaviors in Bakeman (1997). Here we will simply call this an "event" and track its occurrence over time, for two agents, as shown in Figure 1.



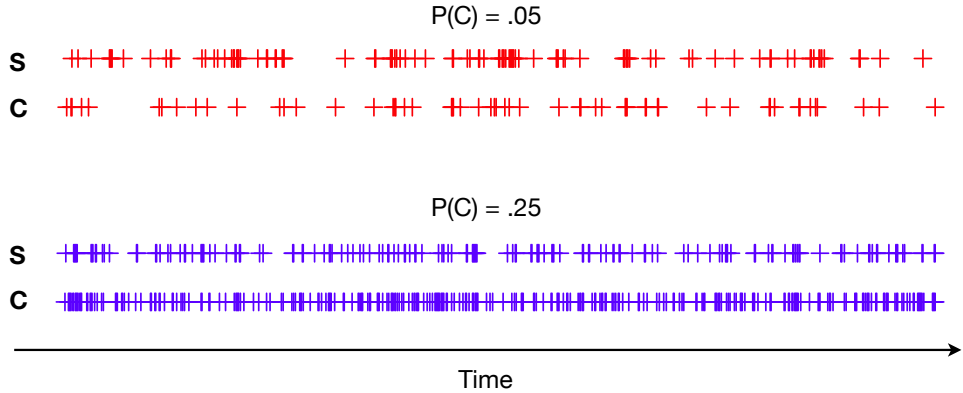

*Figure 1.* Two example experimental runs, in which we observe the behavior of two simple conversational "agents," a conference (C) and participant (S), over 1,000 time steps. The confederate's behavior is experimentally setup to amplify the occurrence of the event. P(C) in the plot reflects the raw probability that the confederate will emit the behavioral event (see Table 1). As specified in the agent's policies, an increase in the behavioral event by the confederate should also increase it in the participant, which are analyses are meant to bear out.

The raw data that this study would use, presumably, is a proportion, aggregated over time, of the behavioral event of interest. In Figure 2, one can see that these events are then aggregated into one rate score. The left side of the plot shows a relatively higher incidence of the behavioral event by the participant agents, compared to the right side of the plot. In our simplified conversational agents, this is a result of the fact that the confederate agents can drive the probability of the event of interest in the participant agent.

Another way of achieving this distinction between low and high conditions is to observe the *correlation* between their behavior and that of the confederate. This is shown in Figure 3, which displays the instantaneous Pearson correlation between interlocutors at different time lags. Such a cross-correlation function gives a more detailed picture of the temporal interaction between interlocutors. The maximal correlation ($\approx$ .2) occurs at a lag of -1, which reflects the confederate agent leading the participant agent.[2] Because a higher event occurrence P(C) generates more events in agent S, the variance accounted for at that lag will also significantly increase, as more events in the confederate will be the driver of those in the participant. Recent exciting extensions of this technique can use a windowed approach to visualize and explore temporal relations, as shown by Boker, Xu, Rotondo, and King (2002) and Barbosa, Déchaine, Vatikiotis-Bateson, and Yehia (2012). In general, cross-correlation informs about the relative covariation between event sequences (i.e., coupling), and its maximal point (in our example, C leads maximally S at lag -1).

This correlation measure has some similarities to aggregation, as they can be described as 'co-aggregation' – observing how the rate of a behavior covaries with that of another time series. Co-variation methods of this kind are obviously useful and fruitfully applied in many contexts, but even

---

[2]Note that the side on which this lag occurs is chosen by the experimenter, and simply reflects the fact that *one* system is leading the other in some way.



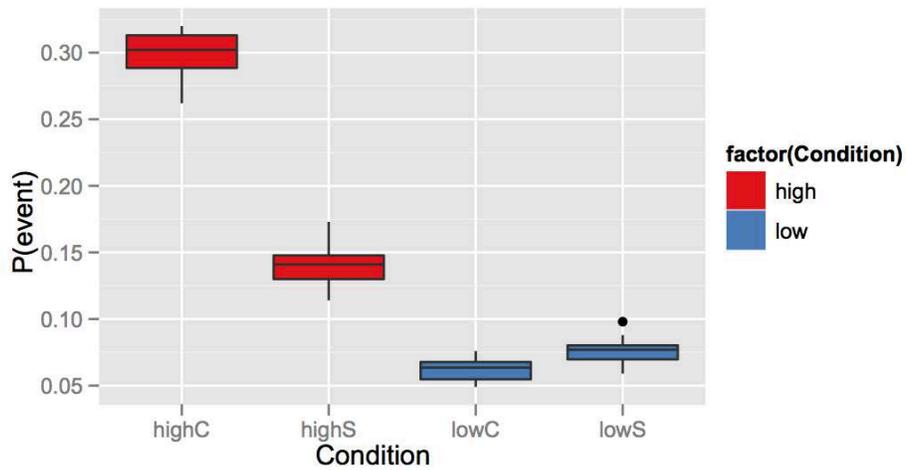

*Figure 2*. Data from 20 simulated interactions for each condition of the confederate's event occurrence rate (.05 vs. .25). As expected, one sees a relative increase in the event's occurrence in agent S if it occurs in agent C.

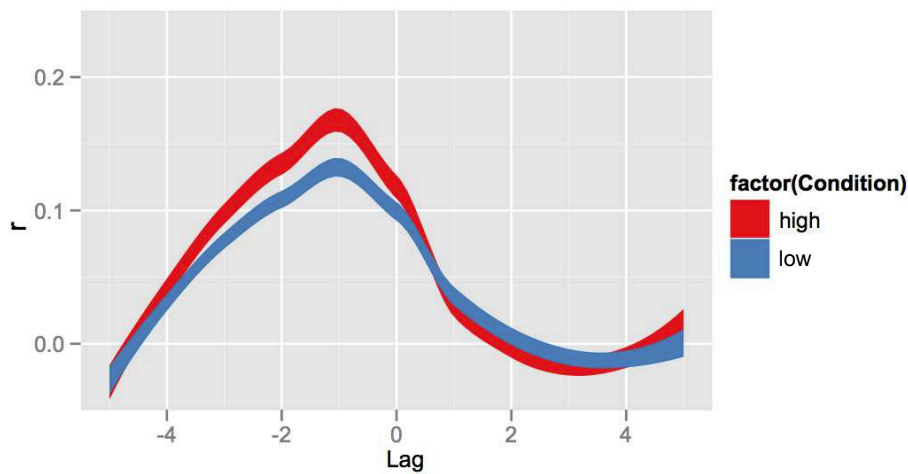

*Figure 3*. Unfolding aggregate scores using cross correlation. Cross-correlation functions between confederate and participant agents. The high agent condition (red), reflecting the cross-correlation between C and S agents at different time lags or shifts (scale: step increments), shows maximal variance accounted for at lag -1, C leading S by one time step (as set in the simulation).



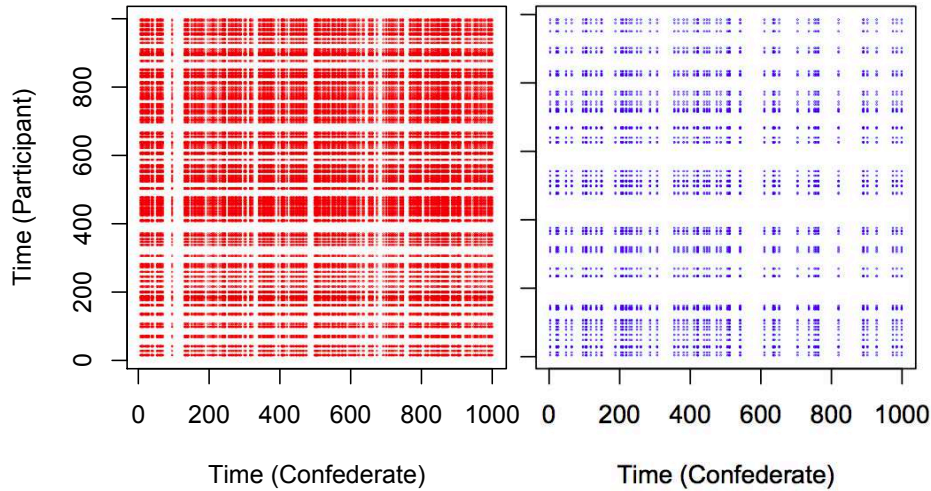

*Figure 4.* Example cross recurrence plots (CRPs) of two sample runs of the simulated data. Left shows a high condition run, right shows a low condition runs. Points reflect relatively in time where C and S are revisiting event states (=1), whereas 0's (nonevents) do not produce points on the plot. The main diagonals reflect lines of coincidence (LOCs), and the lags over which rates are calculated are highlighted.

beyond correlation there are many temporal patterns worthy of exploring. In the cross-recurrence case, one may be said to be exploring co-visitation patterns: How one time series is *revisiting* states that the other time series has visited. This works by quantifying the pattern of visitation of the two systems, rather than simply quantifying their relative rate of occurrence. First, imagine plotting all points $(i_C, j_S)$ where $i_C$ are the *time indices* of the event in agent C's time series, and $j_C$ are the indices of the event in agent S. This produces a visualization of the pattern of co-visitation over time between the two systems. This is shown in Figure 4. These are referred to as cross-recurrence plots (CRPs).

Cross-recurrence quantification analysis (CRQA) is the quantification of the patterns of co-visitation taking place on these plots. Already, one can simply see that there is a much greater density of points on the high condition plot than the low condition. Here we show that quantification of the plots can obtain similar information to cross-correlation, but under a different interpretive scheme. In fact, as we show in the next section, there is a whole range of measures that can be extracted from these plots, and they can become quite sophisticated in their potential implications for the properties of cross recurrence taking place between the two systems that are being compared.

The line of coincidence (LOC) on this plot is where $i_C = j_S$, where the points reflect the systems doing the same thing at the same time. By calculating the *rate* of the event recurrence along the diagonals around the LOC, we obtain a diagonal-wise recurrence rate (RR) measure that also provides a functional characterization of coupling (again, maximized at -1). However, the results will be



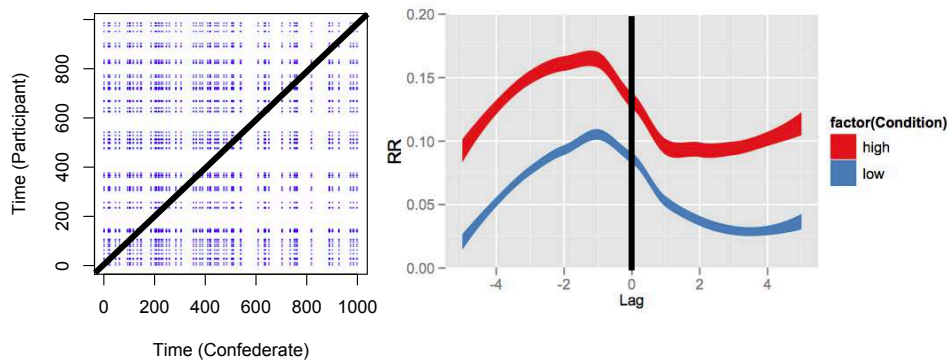

*Figure 5.* By calculating the rate of points on diagonals around the line of coincidence (LOC), we obtain a *diagonal-wise recurrence* that reflects the relative co-visitation, as a function of lag. Like cross-correlation we get a maximization at -1, reflecting C driving S. However, the difference between these functions is larger, proportional to the relative rate of occurrence.

more directly influenced by the rate of co-visitation, or recurrence. So while cross-correlation gives a general measure of the co-variation between two series, cross recurrence shows a co-visitation score that will vary across experimental conditions.

Though this simple diagonal-wise RR profile, computed along the LOC, correlates with cross-correlation (especially in these simple cases), the overall measures will behave differently depending on occurrence rate. It is also important to note that cross recurrence provides the researcher an option to remove the nonevent matches (0's), whereas in cross-correlation they are preserved and implicitly counted toward co-variation (for discussion see Dale, Warlaumont, & Richardson, 2011)[3]. Below we go beyond this simple diagonal-wise RR measure, showing that CRQA also affords an array of other measures to characterize coupling between time series. And in fact, most of these other measures have no obvious analog with the cross-correlation function. These properties have led some to refer to CRQA as a "generalisation of the linear cross-correlation function" (Marwan et al., 2007, p. 256).

Here we have used a simple toy system to compare and contrast aggregation, co-variation, and co-visitation analyses. If one is simply interested in raw rates of occurrence, then aggregation is adequate. However if the researcher wishes to explore functional relationships between systems, cross-correlation or cross-recurrence methods may shed detailed temporal light on their relationship. Cross-correlation measures aggregate co-variation between the two systems, and the maximal correlation observed reflects a stable coupling function between the two systems. However it does not preserve relative rate of "co-visitation" of event states by the two systems. A similar source of information about coupling can be obtained by calculating diagonal-wise RR from cross-recurrence plots, providing both a coupling function *and* a relative rate of occurrence of one system visiting the

---

[3]It is worth noting here that in practice, the 0 event codes are recoded differently for two time series, as distinct "non-event" codes, such as 11 or 12 (for example) to make sure that these non-events do not produce recurrence points on the plot



events of another. As just noted, this is just one simple measure among many provided by CRQA.

Now that we have, it is hoped, motivated the basic interpretive frameworks afforded by these analyses, we delve into CRQA in the next sections and detail how to use the R library. First, we provide more formal details about CRQA and the way it is computed, then explain the most important functions implemented in the `crqa` library and briefly describe the data available to test it. Finally, we compare the computational accuracy and efficiency of our R package with the state of the art MATLAB toolbox, `crptoolbox` (version 5.15) by Marwan et al. (2007) on simulated dichotomous time series, generated as for Table 1. We report tests on: 1) computational *efficiency* (user elapsed time) of the libraries as a function of the length of the time-series and 2) *consistency* (absolute difference and correlations) of the recurrence measures, e.g., recurrence rate, obtained by the two libraries.

## Principles of CRQA

As sketched in the last section, cross-recurrence quantification analysis has been developed to capture the recurring properties and patterns of a dynamical system, which results from two streams of information interacting over time (Zbilut, Giuliani, & Webber, 1998). In behavioral sciences, such streams of information can either be as 'concrete' as body sways or eye-movement trajectories (Shockley et al., 2003; D. C. Richardson & Dale, 2005), but they can also be more 'abstract' sequences of linguistic information, such as the words exchanged by two interlocutors during a dialogue (Fusaroli et al., 2012).

CRQA may thus shed light on the information-feedback dynamics occurring while actions (non-linguistic, linguistic) are transmitted, received, and responded to incrementally by participants in dialogue. So, in the context of a communicative task, CRQA quantifies, for example, how much delay is needed for a listener to be maximally aligned to the instruction delivered by the speaker, how much alignment is observed overall, and so on. In this section, we flesh out these details of the analytic framework, and offer further demonstration of how they work.

Usually CRQA is explained by reference to concepts from dynamical systems. We assume to have measured a time series – one measurement sampled over time – from two systems. Though this single measurement is probably a one-dimensional scalar, CRQA starts by overlaying delayed copies of this time series, for each system separately (displayed in the top row of Figure 6, illustrating this process for one time series). CRQA compares two time series by calculating the degree of their recurrence when these delays are introduced with different numbers of copies, or "embedding dimensions." Specifically, from an original time series $X(t)$, delayed copies $X(t + \tau)$ are generated by introducing a lag $\tau$ into the original time series. The different dimensions of embedding are obtained by considering multiple lags $X(t + m\tau)$.

If in 2 or 3 dimensions, we can plot this delay/copy process, as shown in the bottom-left of Figure 6. This is often referred to as a system's "reconstructed phase space." The phase space consists of the different intervals over which the delays are assigned. We can carry out what is known as "autorecurrence analysis" on this single time series, as shown in the bottom-right recurrence plot in Figure 6. From the plot, measures are based on the number of contiguous points, aligned along the diagonals or along the vertical lines. These lines reflect how the system is revisiting regions of its reconstructed phase space, and points are drawn on the plot when the system is within a certain threshold (illustrated by the circle in Figure 6). "Cross" recurrence uses precisely this process of delaying and embedding, but it is done with two time series. In other words, we reconstruct the



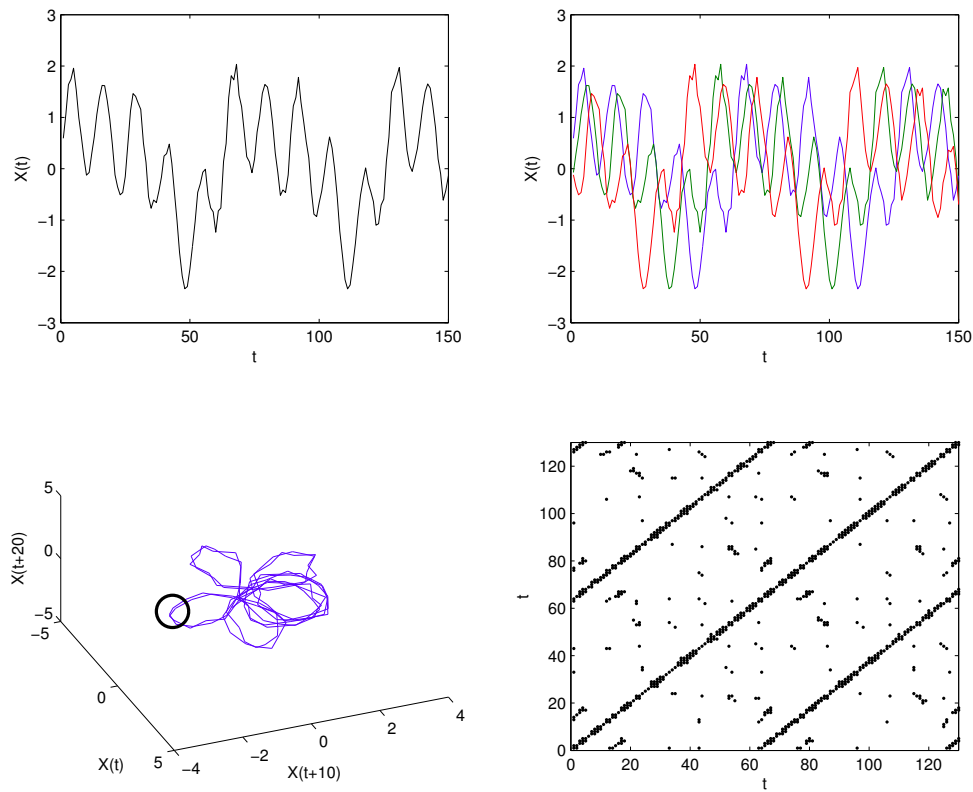

*Figure 6.* A basic sketch of how recurrence is constructed from one time series (top left). The time series is lagged (by 10), copied (3 times), and overlaid with itself (top right). If we use 3 dimensions (copies), then it is possible to visualize this reconstructed phase space (bottom left). By drawing a radius of a given size around parts of this reconstructed phase space (thick line, bottom left), one can determine when recurrence is taking place. The time indices of these recurrence points can be used to constructed the recurrence plot (bottom right). Cross recurrence is done in almost exactly the same way, except two time series are used.

phase space for two time series separately, then see where each respective series' trajectories are nearing each other.

This more complex process is most meaningful in the *continuous* case. A visualization of this is shown in Figure 6. As seen here, a continuous signal is being projected into a higher-dimensional space by taking delayed copies of itself. This can also be done with two time series, and observing where these co-visit each other. Typically researchers set a threshold for determining whether the proximity between the time series is "recurrence" (visualized as a 3D sphere in Figure 6). Helpful best practices can be found in Webber Jr. and Zbilut (2005).

In the previous section we did *precisely* this for the event series, but in quite a simple way. The embedding dimension was set to 1, which essentially projects the event series into the same (one) dimension. In addition, we set a threshold to 0, meaning that an event had to match. Though we extracted *RR* measures across the diagonals, here we describe that many measures can be computed



from these plots, and showcase how the `R` library does it. These measures are derived from the patterns on the plot, often in the form of the diagonal lines reflecting sequences of revisited trajectory regions.

From diagonal lines the measures that are implemented in our `crqa` package are: 1) recurrence rate (*RR*), which is the density of recurrence points in a recurrence plot; 2) percentage determinism (*DET*), which is the percentage of recurrence points forming diagonal lines in the recurrence plot given a minimal length threshold; 3) the length of the longest diagonal ($L_{max}$); 4) the average of the diagonal length (*L*); and 5) the entropy of the diagonal lines (*ENTR*). From the vertical lines, two more mores can be derived: 6) laminarity (*LAM*) the percentage of recurrence points which form vertical lines, and trapping-time (*TT*), which is the mean length of vertical lines. Formal definitions of these measures can be found in Marwan et al. (2007) and on the extremely comprehensive web resource http://www.recurrence-plot.tk/ (by Marwan).

As noted, CRQA can be computed on categorical, as well as on continuous-valued time series. In the categorical case, e.g., a sequence of words, a point recurs when the two series share the same state (i.e., the same word) at two points in time. Recurrence, in this case, can be obtained by means of contingency tables, making cross-recurrence analysis equivalent to lag sequential analysis (Dale, Warlaumont, & Richardson, 2011). At each lag $\tau$, a contingency matrix *CT* is constructed, where each element of it represents the number of times the pair of objects $(i, j)$ co-occurs between the two scan patterns *x* and *y*. More formally: $CT_{i,j}(\tau) = \sum_{t=1}^{t=T-\tau} q(t)$, where *T* is the length of each scan pattern and $q(t) = 1$ if $x(t) = i$ and $y(t + \tau) = j$, and $q(t) = 0$ otherwise. So, if interlocutor C is uttering the word *cat*, and interlocutor S is instead uttering the word *dog*, we fill the *CT* at the corresponding $i, j$ position. From *CT*, recurrence *RR* is computed along the diagonal of *CT* by adding the frequencies of looks to the same objects. Obviously a *CT* has the advantage of measuring co-occurrences between all objects at every lag, making it possible to track how different word co-occurrence contributes to recurrence. The function `CTcrqa`, explained in the next section, is used to compute recurrence of categorical time series by means of *CT* (for more discussion of lag sequential analysis see Bakeman, 1997; Bakeman & Quera, 2011).

As already indicated, continuous measures, such as body sway or acoustic speech energy, may also be subjected to cross-recurrence methods. A distance between points is calculated (our library implements the simplest *Euclidean* distance case), and two points are considered as recurring if they fall within a certain *radius*. When dealing with continuous information, in fact, recurrence cannot be calculated just by looking at the match/mismatch between states for every lag, as distances between points results in continuous values. Thus, the additional step involves the evaluation of a radius, which is a threshold constant used to define whether the distance between points is sufficiently small to consider the two points as recurrent. Setting up an optimal radius is not an easy task, as it strongly depends on the type of dataset analyzed. In section we present a function `optimizeParam` that applies principles of phase-space reconstruction (Marwan et al., 2007) to determine the "optimal" value for radius and other parameters involved in the calculation of cross-recurrence. In Figure 7, we show two examples of categorical and continuous time series extracted from eye-movement information collected during a dialogue experiment conducted by (Coco, Dale, & Keller, n.d.), and provides a visualization of the concept of radius.

These and other data samples are provided in the `crqa` package to illustrate the use and to allow a practical hands-on the main functions. In particular, the `crqa` package (`data(crqa)`) contains:

- `RDts1`, `RDts2`: Two categorical eye-movement series (speaker and listener) of fixated ob-



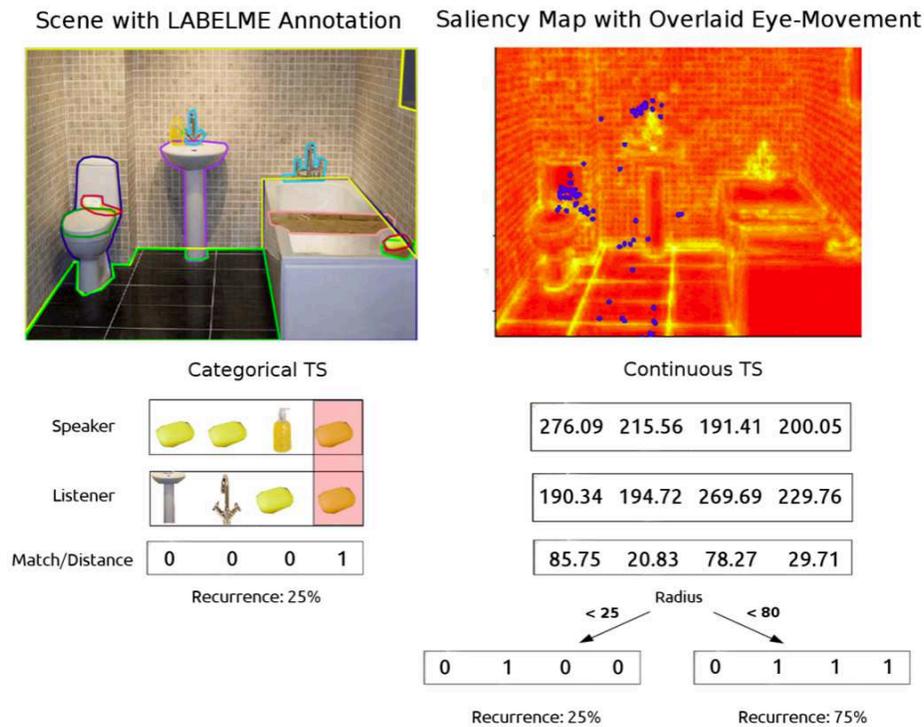

*Figure 7*. Data available in `crqa`. Eye-movement responses of dyads (speakers and listeners) engaged in a spot-the-difference game. Each scene was annotated with polygons using LABELME (Russell, Torralba, Murphy, & Freeman, 2008); which were used to map x-y coordinates of fixations into categorical sequences of fixated objects (left panel, `catts1`, `catts2`). Moreover, we computed visual saliency map of the scene (Torralba, Oliva, Castelhano, & Henderson, 2006), and extracted the saliency value at fixation location (right panel). This gave us continuous time-series of visual saliency value at fixation (`contts1`, `contts2`). In the bottom row of the figure, we illustrate the concept of recurrence in categorical and continuous time-series, and the role played by the radius parameter.

jects taken from D. C. Richardson and Dale (2005).

• `catts1`, `catts2`: Another two categorical eye-movement series (speaker and listener) of fixated objects taken from Coco et al. (n.d.).

• `contts1`, `contts2`: Two continuous eye-movement series (speaker and listener) of visual saliency values at fixation location, taken from Coco et al. (n.d.).

The package also includes the function `simts.R`, which generates dichotomous time-series, as explained in Table 1, that can be used to further learn the use of the `crqa` package.

### Functions

The library provides the user with two methods of computing cross-recurrence between two time series. First, it includes a faster and simpler calculation of only the diagonal-wise recurrence profile, as demonstrated in the section motivating recurrence above, which contains information both about



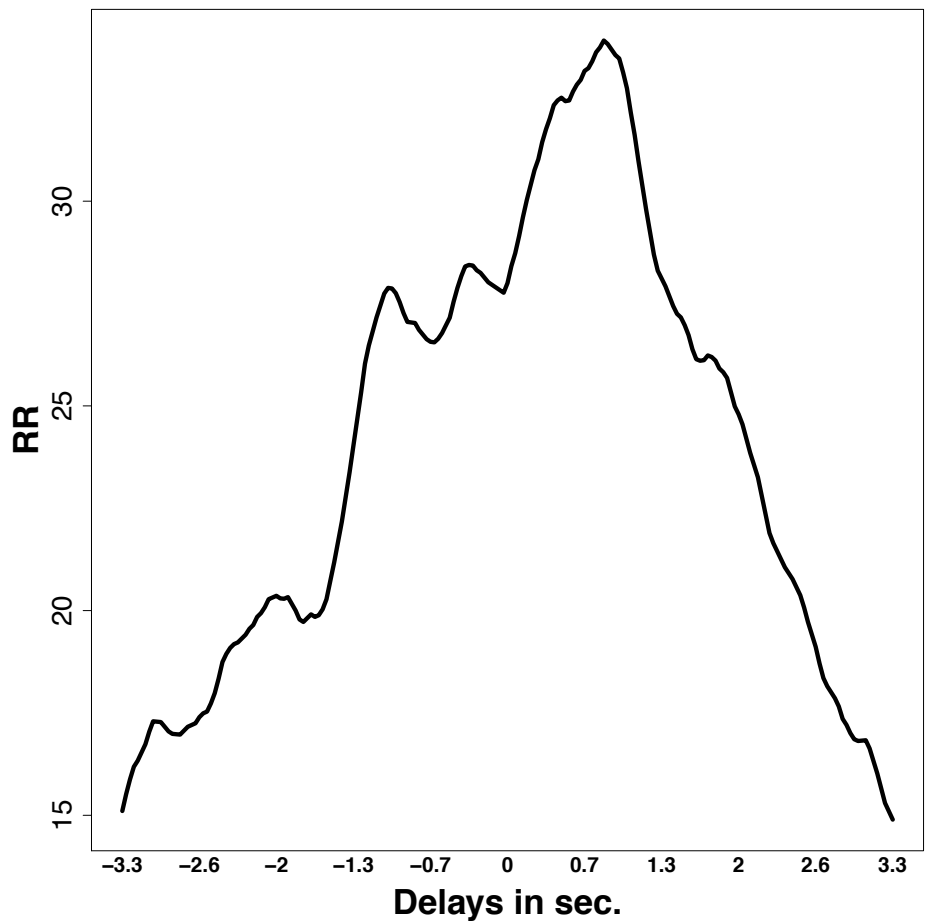

*Figure 8.* Diagonalwise recurrence profile for two eye-movement series (RDts1, RDts2) taken from
D. C. Richardson and Dale (2005)

relative co-visitation and coupling. The library also includes a second, more detailed method, where
a cross-recurrence plot is built for all possible lags, across all states, and several measures of cross-
recurrence, e.g., percentage determinism, are extracted. Put simply, this second approach extracts
all common CRQA measures.

To compute only the diagonal-wise recurrence profile of the two series, we implemented two
functions: `drpdfromts` and `windowdrp`. The function `drpdfromts` implements a quick method to
extract, and explore, the diagonal-wise recurrence profile of two time series. It returns the recurrence
observed for different delays, the maximal recurrence observed, and the delay at which it occurred
(as demonstrated in the section above).

In Figure 8, we show the diagonal-wise recurrence profile for the two series `RDts1, RDts2`.
Each time series is 2,000 datapoints and are from one pair of a speaker and a listener, respectively,
of the dialogue dataset by D. C. Richardson and Dale (2005). The recurrence profile has the typical
leader-follower pattern, where the follower needs a lag of a couple of seconds to be maximally
aligned with the speaker's eye movements.



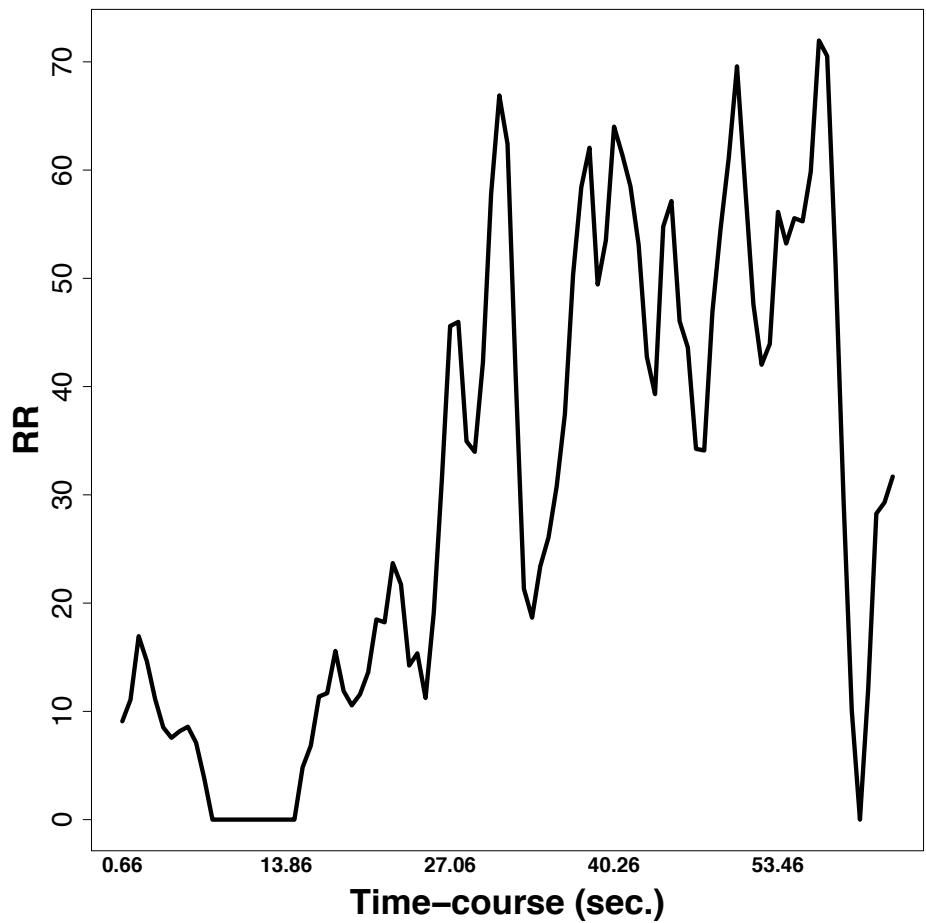

*Figure 9.* Window cross-recurrence of the two eye-movement series (RDts1, RDts2) from
D. C. Richardson and Dale (2005)

When using `drpdfromts`, for categorical sequences, the radius should be set to a very small
value (near 0, e.g., .001). As the categories in the sequence (e.g., *soap*) are recoded into numbers
(e.g. 1), setting the radius to very small value would make only the distance between the same cate-
gory, i.e., 0, be accepted. By changing the `datatype` argument to "continuous", the function would
compute cross-recurrence between time series of continuous data, so the series will be maintained
as numerics. Also for continuous data, we would need a value for the argument `radius`. However,
the value of the radius would have to be tailored to the data observed, because each dataset has its
own idiosyncratic properties, e.g., body movement vs. eye movements. Below, we discuss this issue
further, namely choosing starting parameter values for continuous data. We show an early alpha
version of a function that can perform an optimization routine to estimate these parameters, based
on phase-reconstruction principles (Marwan et al., 2007) (see function `optimizeParam`).

The function `windowdrp`, instead, has similarity to windowed cross-correlation analysis as
in Boker et al. (2002), and tracks how cross-recurrence values evolve over the time course. In par-
ticular, CRQA measures are calculated in overlapping windows of a specified size for a number of



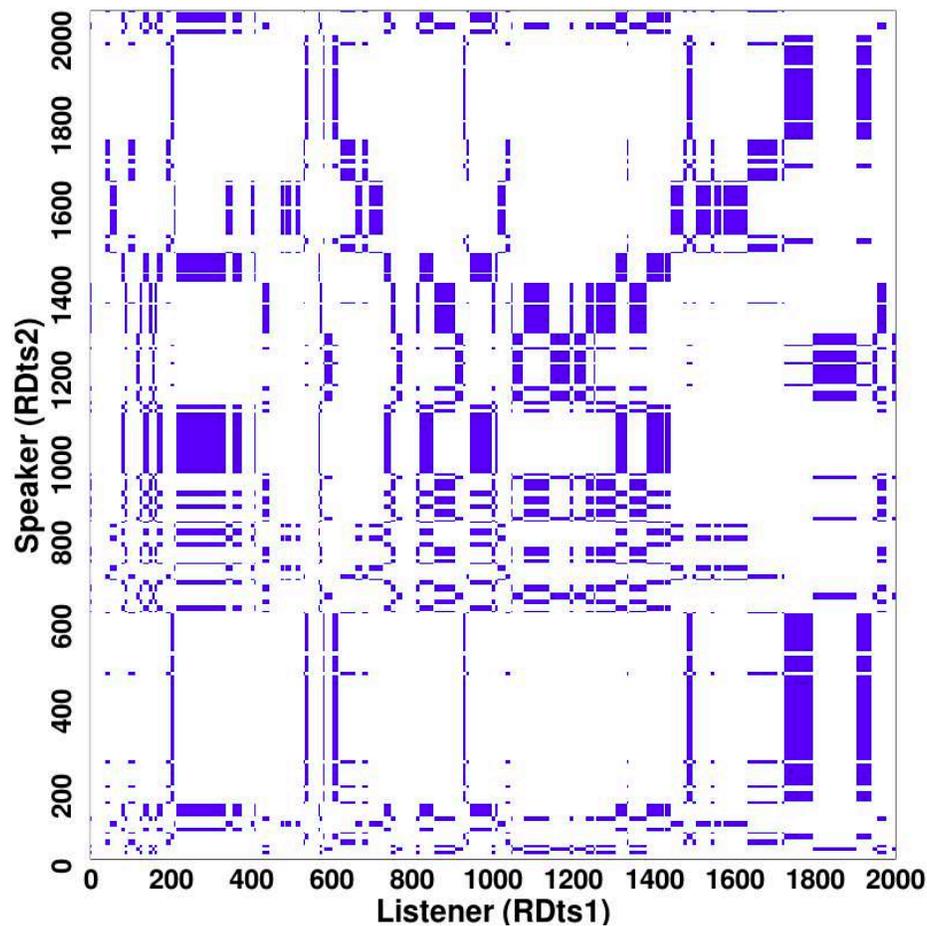

*Figure 10.* Recurrence Plot of the two eye-movement series (RDts1, RDts2) from D. C. Richardson and Dale (2005). The recurrent points are marked with blue color, whereas the non-recurrent points are left blank. The values obtained on the measures for this plot are: $REC = 12.52$; $DET = 98.95$; $Lmax = 124$; $L = 11.3$; $ENTR = 0.66$; $LAM = 97.31$; $TT = 21.32$

delays smaller than the size of the window. In every window, the recurrence value for the different delays is calculated. A mean is then taken across the delays to obtain a recurrence value in that particular window. Tracking recurrence over the time course helps us establishing how the agreement between the two interlocutors develops, as the interaction progresses. We reuse the eye-movement categorical responses `RDts1`, `RDts2`, to display how windowed cross-recurrence between a speaker and a listener evolves as a function of time.

In Figure 9, we can see that about half the time course, the amount of overall recurrence increases, and then fluctuates around the same value till almost the end where it drops. The dyads became more coupled, then recurrence quickly drops as the speaker concludes. Also `windowdrp` can be applied to continuous data by setting up the appropriate `datatype` and `radius` argument, as just described.

More detailed measures characterizing the cross-recurrence of the two time series can be



obtained by using `crqa`. `crqa` is the core function of the package, and examines recurrent structures between time series, which are time-delayed and embedded in higher dimensional space. The approach compares the phase space trajectories of two time-series in the same phase-space when delays are introduced. A Euclidean distance matrix between the two series, delayed and embedded is calculated [4]. On the distance matrix, a recurrence plot is derived by taking all points below a certain radius threshold as recurrent. Several measures representative of the interaction, e.g., recurrence rate (RR), are then extracted (as explained in Principles, above).

In Figure 10, we show the cross-recurrence plot obtained using the two-time series (RDts1, RDts2) from D. C. Richardson and Dale (2005). On the diagonal lines, we observe the pattern of interaction between the two series. The measures characterizing it are *RR*, percentage determinism (*DET*), average and maximal diagonal length (*L* and $L_{max}$), and entropy are calculated. On the vertical lines, we observe the stability of the two series, and relative independence of recurrence over a particular state. The measures characterizing this information are laminarity and trapping-time.

A challenging aspect of computing CRQA is finding appropriate values for the three parameters `radius`, `delay`, `embed`, especially when dealing with continuous time series. The function `optimizeParam` implements an iterative procedure that in three steps attempts to find such values. In particular, the function first identifies a delay that accommodates both time series by finding the local minimum where mutual information between them drops, and starts to level off (Shockley, 2005; Marwan et al., 2007). When one time series has a considerably longer delay than the other, the function selects the longer delay of the two to ensure that new information is gained for both. When the delays are close to each other, the function computes the mean of the two delays. Then, as a second step, the function determines the optimal number of embedding dimensions by using false nearest neighbors, and checking when it bottoms out (i.e., there is no gain in adding more dimensions). If the embedding dimensions for the two time series are different, the algorithm selects the higher embedding dimension of the two to make sure that both time series are sufficiently unfolded. Finally, it determines the radius to use for recurrence by selecting the first radius that yields 1-5% RR. Applied on the continuous visual saliency scan-pattern (`contts1`, `contts2`) taken from the dataset of Coco et al. (n.d.), we obtain: *radius* = 2.59, *embedding dimension* = 20, *delay* = 5. Obviously, this procedure should be iterated over a consistent sample of the data, such that a more precise estimate for the values of the parameters can be obtained.

The `crqa` package also provides the user with a wrapper, `runcrqa.R`, which calls all the methods implemented, such as the simple profile recurrence (`drpdfromts`) or the more extensive analysis of the cross-recurrence plot (`crqa`) both when delays are introduced (`method = 'profile'`) and for a time-course analysis of recurrence (`method = 'window'`). The different methods are called using a list `par` of arguments, according to the type of analysis to be performed (refer to the box R code 5 for more details about the arguments and output).

The last function described in this paper is `CTcrqa`, which is used to compute cross-recurrence on categorical sequences by means of contingency tables (Dale, Warlaumont, & Richardson, 2011; Bakeman, 1997). First, it finds the common states, or categories, shared by the two time series, then it builds up a contingency table (CT) counting the co-occurrences of different sets of states between the two series. For example, in Richardson and Dale (2005) 6 possible characters could be fixated on the visual array during the task. These are nominally coded 1-6.

---

[4] The current version of the package only implements the Euclidean distance, but other metrics can be used.



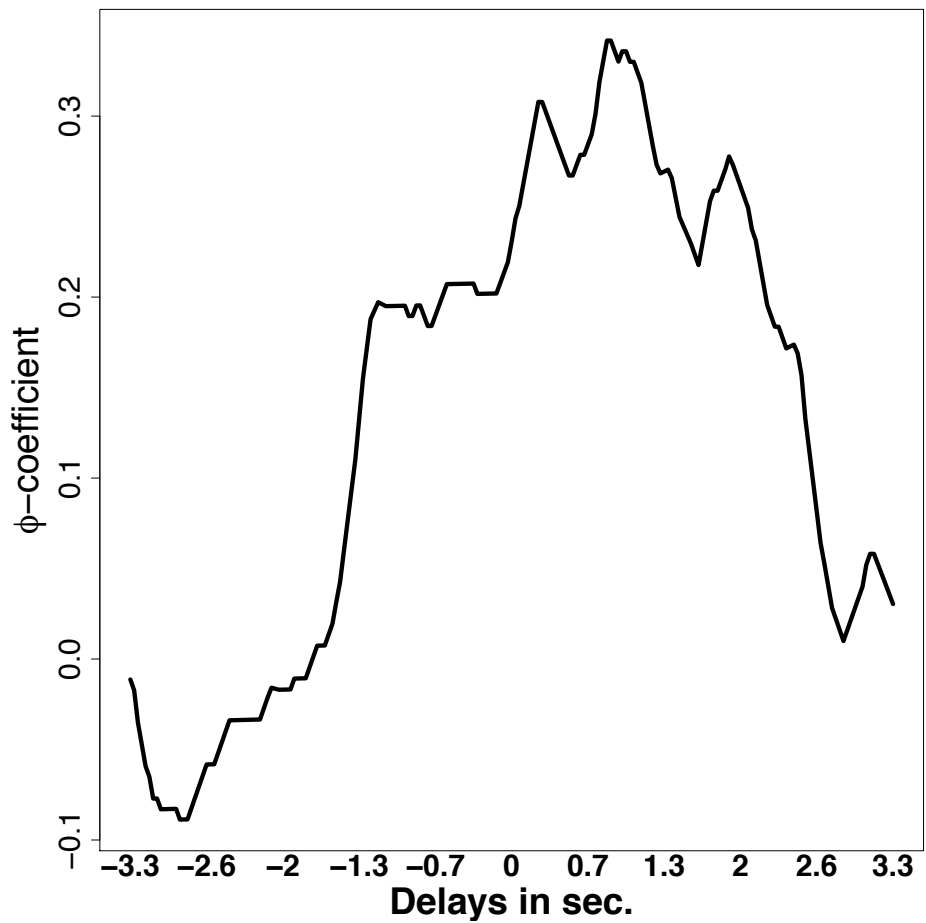

*Figure 11.* Phi-coefficient plot of a particular object for the two eye-movement series (RDts1, RDts2) from D. C. Richardson and Dale (2005).

This contingency-table approach builds a 6x6 table, the cells of which count the number of times speaker/listener were looking at the characters corresponding to that row/column for a given portion of the time series (or, alternatively, the entire time series).

The diagonal of the CT is where the recurrence profile is calculated, as along the diagonal, the states are identical. The advantage of this method is to be able to track co-occurrences of all states involved for each delay introduced. Such values could be potentially used to estimate probability distribution of co-occurrences between states of the two series analyzed, drawing bridges to other sophisticated analytic frameworks, such as lag-sequential analysis (Bakeman, 1997).

When computing recurrence between categorical sequences, we might be specifically interested in a certain object or state. In an eye-tracking dialogue experiment, for example, we might be interested in how looks to a specific target object recur between speakers and listeners. Likewise, in the speech produced by the dyads as they interact, we might be interested in the usage of a specific word referring to that object. The function `calcphi` precisely calculates how recurrence on a specific object between two-series changes when the series are delayed. In particular, the phi(k)



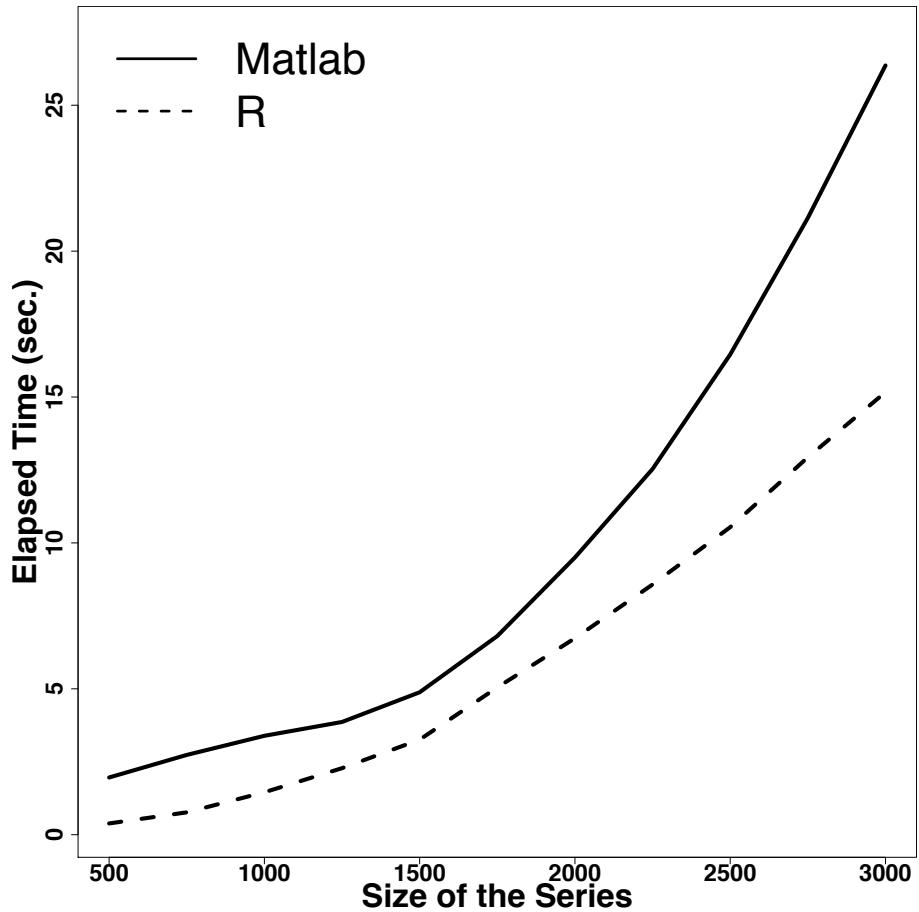

*Figure 12.* Elapsed user time to extract CRQ measures on simulated dichotomous time-series of increasing lengths using `crqa` in R and `crqtoolbox` in MATLAB. Means over 20 iterations are shown as lines. The programming language of the library is identified using line type.

coefficient increases with the frequency of matching recurrence on the same state (k ; k) and away from this state (not k ; not k) between the two time series. On the other hand, phi(k) decreases with the frequency of mismatching objects (k; not-k, and vice versa).

In Figure 11, we show the phi-coefficient for a particular object, coded as 5, looked at in the two series (`RDts1, RDts2`) from D. C. Richardson and Dale (2005). This object was one of six quadrants depicting TV-series characters, that participants had to discuss. In line with Figure 8, we observe the characteristic speaker-leading pattern, whereby the listener takes about one-second to look at object 5, after the speaker has mentioned it.

<div align="center">

**Test of efficiency and consistency**

</div>

We ran 20 iterations and generated two dichotomous time-series with parameter ($P(C)$ = .08, $P(S)$ = .05, $P(C|C)$ = $P(S|S)$ = .05, and $P(S|C)$ = .33, refer to Table 1 for details) of increasing size (from 250 to 3000 ms, steps of 250 ms; 11 different unique size). In a total of 220 simulations, we measure



Table 2

*Linear mixed-effects model predicting* Elapsed Time *as a function of the categorical contrasted predictor* Language *(MATLAB: -0.5, R: 0.5), and the continuous predictor* Size *(a sequence from 500 to 3000 in increases steps of 250 points). Random effect for the model is the Iteration Number.*

| Predictor | β | SE | p |
|---|---|---|---|
| Intercept | -4.853 | 0.379 | 0.0001 |
| Language | 0.4 | 0.536 | 0.4 |
| Size | 0.008 | 0.0002 | 0.0001 |
| Language:Size | -0.002 | 0.0003 | 0.0001 |

the elapsed user time taken to build a CRP and extract from it the following seven measures: *RR* (recurrence rate), *DET* (percentage determinism), $L_{max}$ (length of longest diagonal line), *L* (average diagonal length), *ENTR* (entropy of diagonal lengths above line cutoff, $min > 2$), *LAM* (laminarity of vertical lines) and *TT* (trapping time). For each of the measures, normalized to range between 0 and 1, we compute mean and standard deviation for the absolute distance between the values obtained by R and MATLAB code. Moreover, in order to assess whether the measures obtained with R and MATLAB account for the same variance in the data, we test for correlation and report *t*, degrees of freedom, and *p*-values. Obviously, both packages are tested on the same dataset of simulated time-series. Simulations using `R` (3.0.1 'Good Sport') and MATLAB (2012a) were run with a standard PC machine equipped with an Intel dual core (32 bit), 2.20 GHz, 2.8 GiB RAM, on a Linux OS (Ubuntu 12.04). When calling `crqa` from the `crqtoolbox` in MATLAB by Marwan (2013), we suppressed GUI and other outputs from being printed (i.e., 'silent','nogui').[5]

In Figure 12, we plot mean elapsed user time (y-axis) as a function of sequence lengths. As expected, both libraries demand more time to finish the computation as the time series get longer. However, the `R` implementation outperforms the MATLAB version for increasing size. The significance of these results is assessed using a linear-mixed effect model, in which we predict elapsed time (the dependent measure) as a function of the *Language* used (R, Matlab) and the *Size* of the series[6] (coefficients of the LME models are reported in Table 2). The linear mixed effect model confirms that as size increases by one unit, the amount of time required to complete the computation increases by 8 ms (main effect of size). Crucially, we also found a significant two-way interaction language X size showing that `R` is about 2 ms faster than MATLAB, for every unit increase of series size.

In Table 3, we report the mean absolute difference of the measures obtained using R and MATLAB over 220 simulations. All measures are significantly correlated between the two libraries indicating that the overall performance is perfectly comparable, however, there are some differences to be discussed. We observe exactly the same results for the measures of recurrence, percentage determinism, maximal and mean diagonal line; where we have an absolute difference of 0 and a correlation of 1. The entropy of the diagonal lines is also highly correlated, i.e., $\rho = 0.73$, even though the measures returned by the two libraries differ. Possibly, this difference emerges as in `crqa` we exclude all diagonal lines smaller than a minimal threshold when we compute entropy,

---

[5]Note, even silencing all outputs, a waiting box was automatically launched, and could not be suppressed

[6]Iterations was the random grouping variable



Table 3

*Mean and standard deviation of the absolute difference between measures of cross-recurrence calculated using the R package* `crqa` *and MATLAB toolbox* `crptoolbox`. *Reported also t-value, degree of freedom, p-value and* ρ *from correlation tests between the measures obtained in R and MATLAB*

| Measure | Mean | SD | *t* | *df* | *p* | ρ |
|---------|------|-----|-----|------|-----|-----|
| REC | 0 | 0 | 15478.02 | 218 | 0 | 1 |
| DET | 0 | 0 | 14995.20 | 218 | 0 | 1 |
| Lmax | 0 | 0 | Inf | 218 | 0 | 1 |
| L | 0 | 0 | 11163.43 | 218 | 0 | 1 |
| ENTR | 0.46 | 0.15 | 15.89 | 218 | 0 | 0.73 |
| LAM | 0.07 | 0.03 | 3.54 | 218 | 0 | 0.23 |
| TT | 0.11 | 0.06 | 7.15 | 218 | 0 | 0.43 |

whereas in `crptoolbox`, all diagonal lines seem to be taken into account. The other two measures returned by `crqa` are laminarity and trapping times, both based on vertical lines observed in the plot. The results are significantly correlated, however, again we observe differences to emerge. This difference might be due to the use of the Theiler window parameter in the MATLAB `crptoolbox` code, which is not implemented yet in our `R` code.

We have shown that the performance and results obtained with the `crqa` library in `R` are perfectly comparable to the benchmark MATLAB `crptoolbox` toolbox by Marwan (2013). Obviously, the consolidated MATLAB toolbox provides the user with an extremely handy GUI, as well as numerous other functionalities to visualize the results and compute alternative measures from the recurrence plots. In this respect, the MATLAB toolbox by Marwan, et. al., can still be considered the benchmark for recurrence analyses. However, we believe that our library can be expanded in the future to integrate more functionalities; and as R is a free software for statistical computing, such effort would be certainly sustained by its community of committed users.

## General Discussion

Humans are complex systems, dynamically and interactively exchanging information with their surrounding environment. The most prominent manifestation of such dynamism is observed when humans talk with each other, where the behavior of a single individual engaged in the interaction adapts and aligns with the behaviors of the other individuals that are taking part to the interaction (e.g., Pickering & Garrod, 2004).

The alignment occurring between two interacting individuals has been classically quantified using an aggregative approach, i.e., by correlating frequencies of occurrences of a certain behavior (Bargh & Chartrand, 1999). In language science, the aggregative approach has been the most prominent, where alignment has been measured as the number of common linguistic structures (e.g., lexical, syntactic) used by two interlocutors engaged in a communicative task (Brennan & Clark, 1996; Haywood et al., 2005; Branigan, 2007).

However, alignment has an intrinsic temporal structure, as it unfolds over a sender-receiver



feedback mechanism, e.g., turn-taking in dialogue. Such temporal dependence of alignment has been clearly observed taking place on several 'macro' behaviors, such as postural sways (e.g., Shockley et al., 2003; Louwerse et al., 2012), 'micro' behavior, such as eye-movement (e.g., D. C. Richardson & Dale, 2005), as well as, more recently, on language (e.g., Fusaroli et al., 2012).

The statistical modeling approach used to capture how a dynamical system interactively evolves over time is recurrence analysis (Zbilut et al., 1998; Marwan & Kurths, 2002). This approach aims at quantifying the temporal organization of interacting signals by uncovering the *phases* where such signals are recurring, i.e., they are on the same state; and the *delays* over which recurrence develops.

In this paper, we first empirically motivated the crucial difference between correlation (typically used in the aggregative approach), and co-visitation (typically used in the recurrence approach), and demonstrated that the latter offers a distinct analytic framework. Cross-recurrence quantification analysis is an approach to investigate alignment on a large range of behavioral phenomena, quantifying a range of dynamic relationships that hold between two time series. In particular, we generated binary dichotomous time series, where the probability of certain event to occur in one time series is conditioned to the probability that the event will occur in the other time series. In practice, we simulated an extremely simple interactive system, which can resemble statistical characteristics of real behaviors, such as nodding, or smiling. By using cross-recurrence quantification analysis, we demonstrated that we can capture the same patterns of an aggregative approach, and go beyond that by uncovering the temporal phases during which the interaction takes place.

The advantages of cross-recurrence analysis over more classic approaches to the study of dynamical systems, have called the attention of many researchers, across different fields in cognitive science. Such attention is, in fact, reflected by the amount of recently published work, spanning several topics, where cross-recurrence quantification analysis is used (Fusaroli et al., 2012). A comprehensive bibliography using these methods can be found on http://www.recurrence-plot.tk/ (by Marwan). It includes papers in cognitive science, and many other scientific disciplines.

It appears that the most frequently used software to perform this type of analysis is the MATLAB toolbox `crptoolbox` by Marwan (2013). Even though, `crptoolbox` is an excellent tool to perform cross-recurrence analysis, the research community still lacks an efficient open-source `R` library to perform this analysis. In the second part of this paper, we explained more formally the principles of CRQA analysis, and described an open-source `R` package we developed, `crqa`, which provides to a broad audience the tools to carry out cross-recurrence quantification analysis.

Our package contains functions to quantify cross-recurrence at different levels of analyses. In particular, `drpdfromts` constructs diagonal-wise recurrence profiles of the two time-series across different lags, while `windowdrp` returns a windowed cross-recurrence analysis where recurrence is tracked over the time-course. These two functions just look at the overall cross-recurrence shape. `crqa` instead performs a complete analysis of the cross-recurrence plot returning several measures, such as recurrence rate, percentage determinism, etc. characterizing the dynamics of interaction taking place in the system. By using principles of phase-space reconstruction (Marwan et al., 2007), our library also includes an alpha function, `optimizeParam`, to estimate 'optimal' values for the parameters of *radius*, *delay*, and number of *embedding* dimension. Moreover, the library makes available a function to compute cross recurrence analysis on categorical data by means of contingency tables `CTcrqa`. The advantage of this function, yet to be fully exploited, is that it potentially returns a co-occurrence matrix of all states of the two series at each delay. Such co-occurrence statistics might be integrated in future development of the `crqa` to better estimate recurrence properties of categorical



series.

After presenting the most important functions included in our package, we compared its computational efficiency and consistency with the benchmark MATLAB toolbox (`crptoolbox`) developed by Marwan (2013). By using simulated dichotomous time-series, we demonstrated that our library can be computationally more efficient than its MATLAB rival. In particular, we observed that our `R` library maintained a better elapsed user time as a function of increasing set sizes. Besides being computationally efficient, our package returns measures, which are consistent with those generated by `crptoolbox`.

Even though our `crqa` package achieves remarkable performance, it cannot yet substitute the older and proven `crptoolbox` by Marwan (2013). In fact, `crptoolbox` implements a very handy GUI, integrates many functionality for plotting, as well, it includes additional recurrence measures. Thus, our package will complement rather than substitute `crptoolbox`, by providing the open-source alternative for computing cross-recurrence to a community of researchers, who do not have access to licensed products. Moreover, we believe that the functionalities available in the package will expand in the future with the contribution of its community of users.

## References


Bakeman, R. (1997). *Observing interaction: An introduction to sequential analysis*. Cambridge University Press.

Bakeman, R., & Quera, V. (2011). *Sequential analysis and observational methods for the behavioral sciences*. Cambridge University Press.

Balasubramaniam, R., Riley, M., & Turvey, M. (2000). Specificity of postural sway to the demands of a precision task. *Gait & posture*, *11*(1), 12–24.

Barbosa, A., Déchaine, R., Vatikiotis-Bateson, E., & Yehia, H. (2012). Quantifying time-varying coordination of multimodal speech signals using correlation map analysis. *The Journal of the Acoustical Society of America*, *131*, 2162.

Bargh, J., & Chartrand, T. (1999). The unbearable automaticity of being. *American Psychologist*, *54*(7), 462–479.

Beer, R. (2003). The dynamics of active categorical perception in an evolved model agent. *Adaptive Behavior*, *11*(4), 209–243.

Boker, S., Xu, M., Rotondo, J., & King, K. (2002). Windowed cross-correlation and peak picking for the analysis of variability in the association between behavioral time series. *Psychological Methods*, *7*, 338–355.

Branigan, H. (2007). Syntactic priming. *Language and Linguistics Compass*, *1*(1-2), 1–16.

Brennan, S., & Clark, H. (1996). Conceptual pacts and lexical choice in conversation. *Journal of Experimental Psychology: Learning, Memory, and Cognition*, *22*(6), 1482–1493.

Coco, M., Dale, R., & Keller, F. (n.d.). The role of interactivity on cognitive alignment and decision making during dialogue. *Psychological Science*. (submitted)

Dale, R., Kirkham, N. Z., & Richardson, D. C. (2011). The dynamics of reference and shared visual attention. *Frontiers in Psychology*, *2*.

Dale, R., Warlaumont, A. S., & Richardson, D. C. (2011). Nominal cross recurrence as a generalized lag sequential analysis for behavioral streams. *International Journal of Bifurcation and Chaos*, *21*, 1153–1161.

Fusaroli, R., Bahrami, B., Olsen, K., Roepstorff, A., Rees, G., Frith, C., & Tylén, K. (2012). Coming to terms quantifying the benefits of linguistic coordination. *Psychological science*, *23*(8), 931–939.

Haywood, S., Pickering, M., & Branigan, H. (2005). Do speakers avoid ambiguities during dialogue? *Psychological Science*, *16*(5), 362–366.





Louwerse, M., Dale, R., Bard, E., & Jeuniaux, P. (2012). Behavior matching in multimodal communication is synchronized. *Cognitive science*, *36*(8), 1404–1426.

Marwan, N. (2008). A historical review of recurrence plots. *The European Physical Journal Special Topics*, *164*(1), 3–12.

Marwan, N. (2013). *Cross recurrence plot toolbox.* Retrieved from `http://tocsy.pik-potsdam.de/CRPtoolbox`

Marwan, N., Carmen Romano, M., Thiel, M., & Kurths, J. (2007). Recurrence plots for the analysis of complex systems. *Physics Reports*, *438*(5), 237–329.

Marwan, N., & Kurths, J. (2002). Nonlinear analysis of bivariate data with cross recurrence plots. *Physics Letters A*, *302*, 299-307.

Pickering, M., & Garrod, S. (2004). Toward a mechanistic psychology of dialogue. *Behavioral and brain sciences*, *27*(2), 169–189.

Richardson, D., Dale, R., & Kirkham, N. (2007). The art of conversation is coordination common ground and the coupling of eye movements during dialogue. *Psychological science*, *18*(5), 407–413.

Richardson, D. C., & Dale, R. (2005). Looking to understand: The coupling between speakers' and listeners' eye movements and its relationship to discourse comprehension. *Cognitive Science*, *29*, 39–54.

Riley, M., & Van Orden, G. (2005). *Tutorials in contemporary nonlinear methods for the behavioral sciences web book.* Arlington, VA: Digital Publication Available through the National Science Foundation. Retrieved from `http://www.nsf.gov/sbe/bcs/pac/nmbs/nmbs.jsp`

Russell, B., Torralba, A., Murphy, K., & Freeman, W. (2008). Labelme: a database and web-based tool for image annotation. *International Journal of Computer Vision*, *77*(1-3), 151-173.

Schober, M. (1993). Spatial perspective-taking in conversation. *Cognition*, *47*(1), 1–24.

Shockley, K. (2005). Cross recurrence quantification of interpersonal postural activity. In *In M. A. Riley & G. C. Van Orden (Eds.), Tutorials in contemporary nonlinear methods for the behavioral sciences* (pp. 142–177). Digital Publication Available through the National Science Foundation. Retrieved from `http://www.nsf.gov/sbe/bcs/pac/nmbs/nmbs.jsp`

Shockley, K., Santana, M., & Fowler, C. (2003). Mutual interpersonal postural constraints are involved in cooperative conversation. *Journal of Experimental Psychology Human Perception and Performance*, *29*, 326-332.

Shockley, K., & Turvey, M. (2005). Encoding and retrieval during bimanual rhythmic coordination. *Journal of Experimental Psychology: Learning, Memory and Cognition*, *31*(5), 980-990.

Stephen, D., Dixon, J., & Isenhower, R. (2009). Dynamics of representational change: Action, entropy, & cognition. *Journal of Experimental Psychology: Human, Perception and & Performance*, *35*, 1811-1822.

Torralba, A., Oliva, A., Castelhano, M., & Henderson, J. (2006). Contextual guidance of eye movements and attention in real-world scenes: the role of global features in object search. *Psychological review*, *4*(113), 766–786.

Webber Jr., C., & Zbilut, J. (2005). Recurrence quantification analysis of nonlinear dynamical systems. In *In M. A. Riley & G. C. Van Orden (Eds.), Tutorials in contemporary nonlinear methods for the behavioral sciences* (pp. 26–94). Arlington, VA: Digital Publication Available through the National Science Foundation. Retrieved from `http://www.nsf.gov/sbe/bcs/pac/nmbs/nmbs.jsp`

Zbilut, J., Giuliani, A., & Webber, C. (1998). Recurrence quantification analysis and principal components in the detection of short complex signals. *Physics Letters A*, *237*(3), 131–135.




**R code 1: `drpdfromts`**

Extract the cross-recurrence diagonal profile of two-time series.

**Usage**:

```
drpfromts(t1, t2, ws, datatype, radius)
```

**Arguments**:

`t1` = First time-series

`t2` = Second time-series

`ws` = A constant indicating the range of delays (positive and negative) to explore

`datatype` = A string indicating whether the time-series consist of *'categorical'*, or *'continuous'* datatype

`radius` = A threshold, cut-off, constant used to decide whether two points are recurrent or not.

**Output**:

An object list with the following arguments:

`profile` = Recurrence (ranging from 0,1) with length equal to the number of delays explored

`maxrec` = Maximal recurrence observed between the two-series

`maxlag` = Delay at which maximal recurrence is observed

**Example**:

```
res = drpdfromts(catts1, catts2, ws = 40, datatype = ``categorical'',
radius = 0.00001)

profile = res$profile

plot(seq(1,length(profile),1),profile)
```



**R code 2: `windowdrp`**

Window cross-recurrence profile of two time-series, the maximal recurrence observed on it, and its lag.

**Usage**:

```
windowdrp(x, y, step, windowsize, lagwidth, datatype, radius)
```

**Arguments**:
   x = First time-series
   y = Second time-series
   `step` = Interval by which the window is moved.
   `windowsize` = The size of the window.
   `lagwidth` = Delays considered.
   `datatype` = A string indicating whether the time-series consist of *'categorical'*, or *'continuous'* datatype
   `radius` = A threshold, cut-off, constant used to decide whether two points are recurrent or not.

**Output**:

An object list with the following arguments:
   `profile` = Time-course cross-recurrence profile
   `maxrec` = Maximal recurrence observed along the time-series
   `maxlag` = Time-point where maximal recurrence is observed

**Example**:

```
step = 20; windowsize = 100; lagwidth = 40; datatype = "categorical"
radius = 0.00001

ans = windowdrp(catts1, catts2, step, windowsize, lagwidth, datatype
radius)
profile = res$profile

plot(seq(1,length(profile),1),profile)
```



**R code 3: `crqa`**

Core cross recurrence function, which examines recurrent structures between time-series, which are time-delayed and embedded in higher dimensional space.

**Usage**:

```
crqa(data1, data2, delay, embed, rescale, radius, normalize, minline)
```

**Arguments**:

  `ts1` = First time-series.

  `ts2` = Second time-series.

  `delay` = The delay unit by which the series are lagged.

  `embed` = The number of embedding dimension for phase-reconstruction, i.e., the lag intervals.

  `rescale` = Rescale the distance matrix; if rescale = 1 (mean distance of entire matrix); if rescale = 2 (maximum distance of entire matrix).

  `radius` = A threshold, cut-off, constant used to decide whether two points are recurrent or not.

  `normalize` = Normalize the time-series; if normalize = 0 (do nothing); if normalize = 1 (Unit interval); if normalize = 2 (z-score).

  `mindiagline` = A minimum diagonal length of recurrent points. Usually set to 2, as it takes a minimum of two points to define any line.

  `minvertline` = A minimum vertical length of recurrent points.

  `whiteline` = A logical flag to calculate (TRUE) or not (FALSE) empty vertical lines.

  `recpt` = A logical flag indicating whether measures of cross-recurrence are calculated directly from a recurrent plot (TRUE) or not (FALSE).

**Output**:

When CRQA can be calculated, it returns a list with the different measures extracted from the recurrence plot. Otherwise, all output arguments will be either 0 or NA.

  `rec` = The percentage of recurrent points falling within the specified radius (range between 0 and 100)

  `det` = Proportion of recurrent points forming diagonal line structures.

  `nrline` = The total number of lines in the recurrent plot.

  `maxline` = The length of the longest diagonal line segment in the plot, excluding the main diagonal.

  `meanline` = The average length of diagonal line structures.

  `entropy` = Shannon information entropy of all diagonal line lengths.

  `relEntropy` = Entropy measure normalized by the number of lines observed in the plot. Handy to compare across contexts and conditions.

  `lam` = Proportion of recurrent points forming vertical line structures.

  `tt` = The average length of vertical line structures.

**Example**:

```
delay = 1; embed = 1 ; rescale = 1; radius = 0.00001; normalize = 0;
minvertline = 2; mindiagline = 2; whiteline = FALSE; recpt = FALSE

ans = crqa(catts1, catts2, delay, embed, rescale, radius, normalize,
mindiagline, minvertline, whiteline, recpt)

print(ans[1:9])
```



**R code 4: `optimizeParam`**

Iterative procedure exploring a combination of parameter values to obtain values for the three parameters of radius, delay and embedding dimensions that optimize recurrence between two time-series.

**Usage**:

```
optimizeParam(ts1, ts2, par)
```

**Arguments**:
   `ts1` = First time-series.
   `ts2` = Second time-series.
   `par` = A list of parameters for the optimization:
`lgM` = a constant indicating maximum lag to inspect when calculating average mutual information between the two series.
`steps` = a sequence of points (e.g., seq(1, 10, 1)) used to look ahead local minima.
`cut.del` = a sequence of points (e.g., seq(1, 40,1) ) indicating the delays to evaluate when mutual information between the two series is estimated.
**Output**:

It returns a list with the following arguments:
   `radius` = The optimal radius found.
   `emddim` = Number of embedding dimensions
   `delay` = The delay parameter to lag the time-series.
**Example**:

```
par(list = c(lgM = 100, steps = seq(1, 10, 1), cut.del = seq(1, 40,1)))
res = optimizeParam(contts1, contts2, par)

print(res)
```



**R code 5: `runcrqa.R`**

Wrapper to extract CRQ information on categorical and continuous time series informa-
tion. The function provides two types of analysis: the recurrence diagonal profile (type =
1), or a detailed analysis of the recurrence plot (type = 2). For both methods, the function
can perform a profile analysis (method = 'profile'), which looks at how recurrence change
for the different lags, or a window analysis (method = 'window'), where recurrence is
tracked across the time-course by sliding overlapping windows.

**Usage**:

```
runcrqa(ts1, ts2, par)
```

**Arguments**:

Independently of the type, the argument 'method' can take two values either 'profile'
or 'window': `method = 'profile'`: compute the recurrence profile over the all time
series for different lags `method = 'window'`: compute recurrence over time by sliding a
window

   `ts1` = First time-series.

   `ts2` = Second time-series.

   `par` = A list of argument parameters depending on whether the wrapper is used to obtain
only profiles, i.e., type 1, or to extract more detailed measures from the cross-recurrence
plot, type 2. See details below for a detailed explanation of the arguments for the two
different methods.

Please for the details of `par` arguments refer to:

R code 1 (`drpdfromts`) and 2 (`windowdrp`) with:

`type = 1`: with `method = 'profile'` or `method = 'window'`

R code 3 (`crqa`) and `wincrqa` with:

`type = 2`: with `method = 'profile'` or `method = 'window'`

When `method = 'window'` is used, please specify also arguments for
`lagwidth` and `windowsize`.

**Output**:

Also the output values returned depends on the `type = 1|2`, and on the
`method = 'profile'|'window'`. Please refer to previous boxes for details.

**Example**:

Use runcrqa wrapper to calculate only recurrence profile

```
par = list(type = 1, ws = 40, method = "profile",
datatype = "categorical", thrshd = 8)

ans = runcrqa(catts1, catts2, par)

profile = ans$profile; maxrec = ans$maxrec; maxlag = ans$maxlag
```



**R code 6: `CTcrqa`**

Recurrence profile is calculated by means of contingency tables, and it can only be used on categorical time-series.

**Usage**:
```
CTcrqa(ts1, ts2, par)
```

**Arguments**:
    `ts1` = First time-series.
    `ts2` = Second time-series.
    `par` = A list of parameters for the optimization:

`datatype` = a string specifying whether the time-series is 'numerical' or 'categorical'.
`thrshd` = a constant indicating the maximum difference between time-series lengths that is tolerated.
`lags` = a numerical vector for the delays, e.g., seq(1,100, 1)
**Output**:

A cross-recurrence profile of the two-time series with length equal to the number of delays considered.
**Example**:

```
par = list(lags = seq(1, 40, 1), datatype = "categorical", thrshd = 8)

res = CTcrqa(catts1, catts2, par)
```
Show profile
```
plot(seq(1,length(res),1), res, xlab = "Delays",
ylab = "Recurrence", type = "l", lwd = 3)
```



**R code 7: `calcphi`**

Phi-coefficient is the recurrence profile observed between the two time-series on a specific state k.

**Usage**:
```
calcphi(t1, t2, ws, k)
```

**Arguments**:
   `ts1` = First time-series.
   `ts2` = Second time-series.
   `ws` = Number of delays (+/-) considered.
   `k` = The categorical state on which phi is calculated.
**Output**:

It returns the phi-coefficient for state k for all delays considered.
**Example**:

```
k = 5, ws = 100 res = calcphi(catts1, catts2, ws, k)
```